\documentclass[conference]{IEEEtran}
\pdfoutput=1
\IEEEoverridecommandlockouts
\usepackage{cite}
\usepackage{amsmath,amssymb,amsfonts}
\usepackage{graphicx}
\usepackage{textcomp}
\usepackage{xcolor}
\usepackage{bm}
\usepackage{hyperref}
\usepackage{amsthm}
\usepackage{subfigure}
\usepackage[linesnumbered,ruled,noend]{algorithm2e} 
\usepackage{algorithmic}
\usepackage{multirow}
\usepackage{graphicx}
\usepackage{float}
\usepackage{siunitx}
\def\BibTeX{{\rm B\kern-.05em{\sc i\kern-.025em b}\kern-.08em
    T\kern-.1667em\lower.7ex\hbox{E}\kern-.125emX}}
\begin{document}

\title{Attentive Neural Controlled Differential Equations for Time-series Classification and Forecasting}

\author{\IEEEauthorblockN{Sheo Yon Jhin$^1$, Heejoo Shin$^1$, Seoyoung Hong$^1$, Minju Jo$^1$, Solhee Park$^1$, Noseong Park$^1$,\\Seungbeom Lee$^2$, Hwiyoung Maeng$^2$, Seungmin Jeon$^2$}
% \IEEEauthorblockA{\textit{Department of Artificial Intelligence} \\
\textit{Yonsei University$^1$, Seoul, South Korea}\\
\textit{Socar Co. Ltd.$^2$, Seoul, South Korea}\\
\{sheoyonj,xenos961,seoyoungh,alflsowl12,leadwish11,noseong\}@yonsei.ac.kr, \{marcus,colde,simon\}@socar.kr}
% \and
% \IEEEauthorblockN{Heejoo Shin}
% % \IEEEauthorblockA{\textit{dept. name of organization (of Aff.)} \\
% \textit{Yonsei University}\\
% Seoul,South Korea \\
% xenos961@yonsei.ac.kr}
% \and
% \IEEEauthorblockN{Seoyoung Hong
% }
% \IEEEauthorblockA{\textit{Department of Artificial Intelligence} \\
% \textit{Yonsei University}\\
% Seoul,South Korea \\
% seoyoungh.kr@gmail.com}
% \and
% \IEEEauthorblockN{Solhee Park}
% \textit{Yonsei University}\\
% Seoul,South Korea \\
% leadwish11@yonsei.ac.kr}
% \and
% \IEEEauthorblockN{Noseong Park}
% \IEEEauthorblockA{\textit{Department of Artificial Intelligence} \\
% \textit{Yonsei University}\\
% Seoul,South Korea\\
% noseong@yonsei.ac.kr}

\maketitle

\begin{abstract}
Neural networks inspired by differential equations have proliferated for the past several years, of which neural ordinary differential equations (NODEs) and neural controlled differential equations (NCDEs) are two representative examples. In theory, NCDEs exhibit better representation learning capability for time-series data than NODEs. In particular, it is known that NCDEs are suitable for processing irregular time-series data. Whereas NODEs have been successfully extended to adopt attention, methods to integrate attention into NCDEs have not yet been studied. To this end, we present \underline{\textbf{A}}ttentive \underline{\textbf{N}}eural \underline{\textbf{C}}ontrolled \underline{\textbf{D}}ifferential \underline{\textbf{E}}quations (ANCDEs) for time-series classification and forecasting, where dual NCDEs are used: one for generating attention values, and the other for evolving hidden vectors for a downstream machine learning task. We conduct experiments on three real-world time-series datasets and ten baselines. After dropping some values, we also conduct experiments on irregular time-series. Our method consistently shows the best accuracy in all cases by non-trivial margins. Our visualizations also show that the presented attention mechanism works as intended by focusing on crucial information.
\end{abstract}

\begin{IEEEkeywords}
time-series data, neural controlled differential equations, attention
\end{IEEEkeywords}

\section{Introduction}
%Time-series data is one popular data type 
Time-series data is a popular and important data type in the field of data mining and machine learning~\cite{fu2011review, ahmed2010empirical,fawaz2019deep, weigend2018time, esling2012time, kirchgassner2012introduction, krollner2010financial, bontempi2012machine, reinsel2003elements, ralanamahatana2005mining}. From an ordered set of multi-dimensional values, which are tagged with their occurrence time-point information, we typically extract useful information, forecast future values, or classify class labels. 
%For these, many different techniques have been developed, ranging from classical statistical methods to recent deep learning-based methods.
Many different techniques, ranging from classical statistical methods to recent deep learning-based methods, have been developed to tackle this type of data.

Among those deep learning-based methods, recurrent neural networks (RNNs) are considered the standard way to deal with time-series data. Long short-term memory (LSTM) networks and gated recurrent unit (GRU) networks are representative RNNs that are known to be suitable for processing time-series data~\cite{sepp1997long, chung2014empirical, che2018recurrent}.

Owing to the recent advancement of deep learning technology, however, neural ordinary differential equations (NODEs~\cite{NIPS2018_7892}) and neural controlled differential equations (NCDEs~\cite{DBLP:conf/nips/KidgerMFL20}) show state-of-the-art accuracy in many time-series tasks and datasets.

\begin{figure}
    \centering
    \subfigure[NCDE]{\includegraphics[width=\columnwidth]{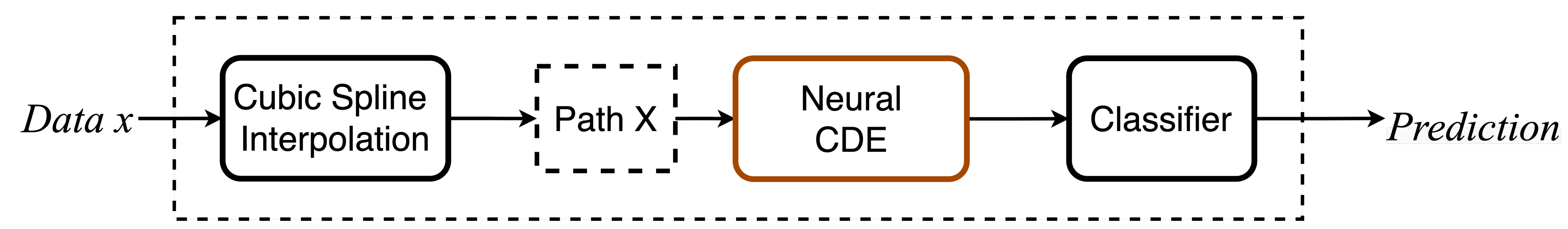}}
    \subfigure[Proposed attentive NCDE]{\includegraphics[width=\columnwidth]{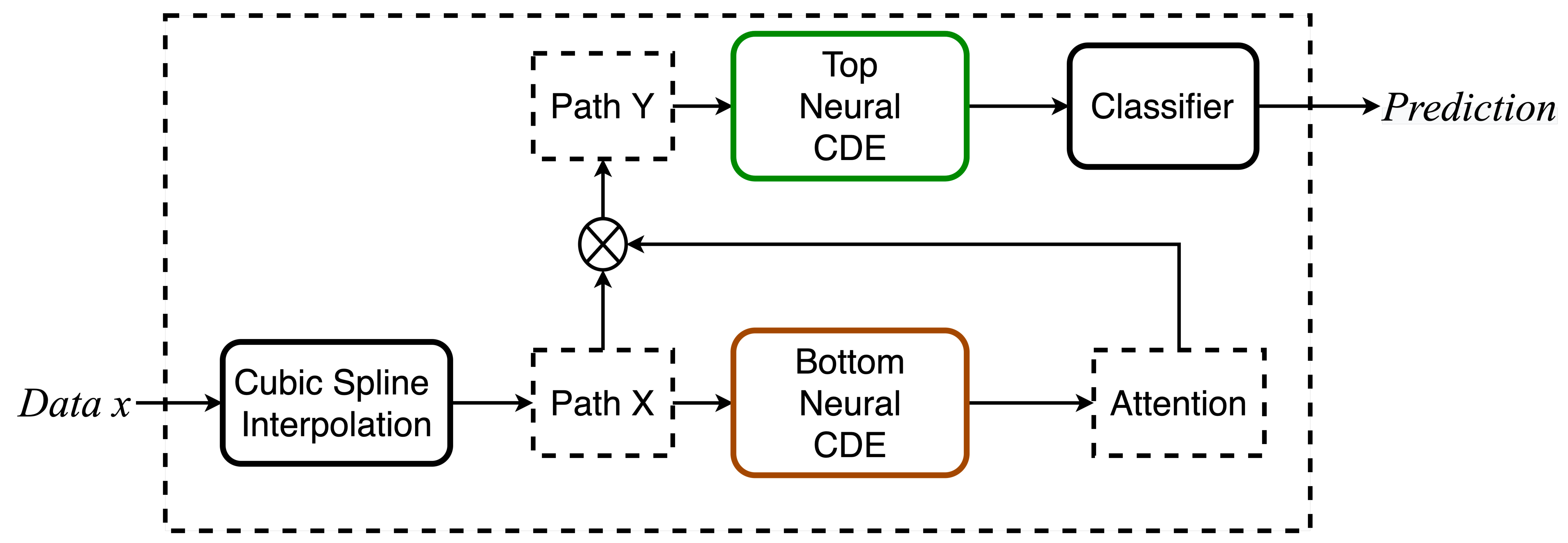}}
    \caption{The overall architecture of NCDE and our proposed attentive NCDE (ANCDE). In our method, the bottom NCDE produces attention values and the top NCDE produces the last hidden vector. This figure assumes classification but we also support regression.}
    \label{fig:archi}
\end{figure}

In NODEs, a multi-variate time-series vector $\bm{z}(t_1)$\footnote{$\bm{z}(t)$ can be either a raw multi-variate time-series vector or a hidden vector created from the raw input.} at time $t_1$ can be calculated from an initial vector $\bm{z}(t_0)$ by $\bm{z}(t_1) = \bm{z}(t_0) + \int_{t_0}^{t_1} f(\bm{z}(t);\bm{\theta}_f) dt$, where $f(\bm{z}(t);\bm{\theta}_f) = \frac{d\bm{z}(t)}{dt}$ and $t_i \in [0,T]$ refers to a time-point. We note that $f$ is a neural network parameterized by $\bm{\theta}_f$, which learns the time-derivative term of $\bm{z}$. Therefore, we can say that $f$ contains the essential evolutionary process information of $\bm{z}$ after being trained. Several typical examples of $\bm{z}$ include but are not limited to climate models, heat diffusion processes, epidemic models, and so forth, to name a few~\cite{debrouwer2019gruodebayes, zang2020neural, portwood2019turbulence}. Whereas RNNs, which directly learn $\bm{z}$, are discrete, NODEs are continuous w.r.t. time (or layer\footnote{In NODEs, the time variable $t$ corresponds to the layer of a neural network. In other words, NODEs have continuous depth.}). Since we can freely control the final integral time, e.g., $t_1$ in the above example, we can learn and predict $\bm{z}$ at any time-point $t$.

In comparison with NODEs, NCDEs assume more complicated evolutionary processes and use $\bm{z}(t_1) = \bm{z}(t_0) + \int_{t_0}^{t_1} f(\bm{z}(t);\bm{\theta}_f) dX(t)$, where the integral is a Riemann–Stieltjes integral~\cite{DBLP:conf/nips/KidgerMFL20}. NCDEs reduce to NODEs if and only if $X(t) = t$, i.e., the identity function w.r.t. time $t$. In other words, NODEs are controlled by time $t$ whereas NCDEs are controlled by another time-series $X(t)$. In this regard, NCDEs can be understood as a generalization of NODEs. %One more thing worth mentioning
Another point worth mentioning is that $f(\bm{z}(t);\bm{\theta}_f) dX(t)$ indicates a matrix-vector multiplication, which can be performed without spending large computational costs. In~\cite{NEURIPS2020_4a5876b4}, it has been demonstrated that NCDEs significantly outperform RNN-type models and NODEs in standard benchmark problems.

%However, NCDEs are still in their early phase since it had been recently developed. 
Despite their promising performance, NCDEs are still a fairly novel concept in their early phase of development. In this work, we greatly enhance the accuracy of NCDEs by adopting \emph{attention}. Attention is a widely-used approach to enhance neural networks by letting them focus on a small set of useful information only, which is also effective in time-series modeling~\cite{10.5555/3295222.3295349}.
%However, it had not been studied how to improve NCDEs with attention and we propose a way to do it.
To the best of our knowledge, no other works have ventured to apply attention to NCDEs.

In our proposed \underline{\textbf{A}}ttentive \underline{\textbf{N}}eural \underline{\textbf{C}}ontrolled \underline{\textbf{D}}ifferential \underline{\textbf{E}}quations (ANCDEs), there are two NCDEs as shown in Fig.~\ref{fig:archi}: one for generating attention values and the other for evolving $\bm{z}(t)$. We support two different attention methods: i) time-wise attention $a(t) 
\in [0,1]$ and ii) element-wise attention $\bm{a}(t) \in [0,1]^D$, where $D = \dim(X(t))$. In both cases, we feed the (element-wise) multiplication of the attention and $X(t)$, denoted by $Y(t)$, into the second NCDE, therefore the second NCDE's input is comprised of values selected from $X(t)$ by the first NCDE. We also propose a training algorithm that guarantees the well-posedness of our training problem.

We conduct experiments on three real-world datasets and tasks: i) hand-written character recognition, ii) mortality classification, and iii) stock price forecasting. We compare our method with RNN-based, NODE-based, and NCDE-based models. In particular, we consider the state-of-the-art NODE model with attention. Our proposed method outperforms them in all cases. Our contributions can be summarized as follows:
\begin{enumerate}
    \item We extend NCDEs by adopting attention.
    \item To this end, we use dual NCDEs. The bottom NCDE produces the attention value of the path $X(t)$. The top NCDE reads the (element-wise) multiplication of the attention and $X(t)$ and produces the last hidden vector for a downstream machine learning task.
    \item We propose a specialized training algorithm that guarantees the well-posedness of our training problem.
    \item Our method, called \emph{ANCDE}, significantly outperforms existing baselines.
    % \item Our source codes and data are in {\color{blue}\url{https://drive.google.com/drive/folders/1Bh39rdPWl}}.
\end{enumerate}

\section{Related Work \& Preliminaries}
In this section, we review NODEs and NCDEs in detail. We also introduce attention.
\begin{figure}
    \centering
    \subfigure[NODE]{\includegraphics[width=\columnwidth]{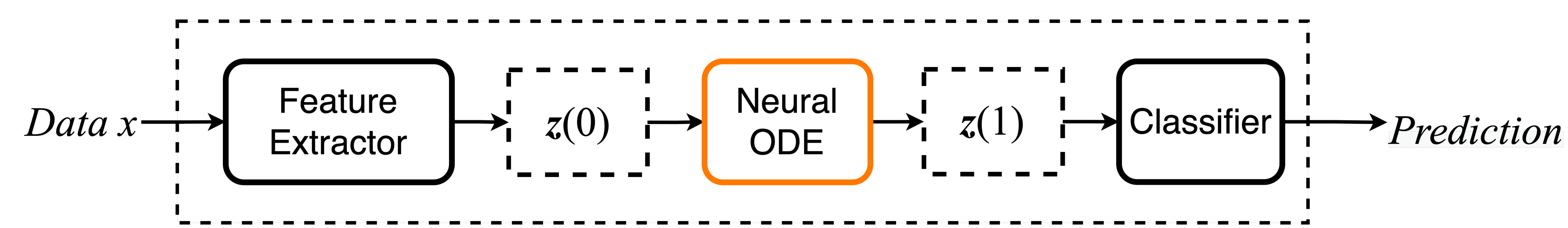}}
    
    \caption{The overall architecture of NODEs}
    \label{fig:archi3}
\end{figure}
\subsection{Neural ODEs}\label{sec:node}
NODEs solve the following initial value problem (IVP), which involves an integral problem, to calculate $\bm{z}(t_1)$ from $\bm{z}(t_0)$~\cite{NIPS2018_7892}:
\begin{align}
    \bm{z}(t_1) = \bm{z}(t_0) + \int_{t_0}^{t_1}f(\bm{z}(t);\bm{\theta}_f)dt,
\end{align}where $f(\bm{z}(t);\bm{\theta}_f)$, which we call \emph{ODE function}, is a neural network to approximate $\dot{\bm{z}} \stackrel{\text{def}}{=} \frac{d \bm{z}(t)}{d t}$. To solve the integral problem, NODEs rely on ODE solvers, such as the explicit Euler method, the Dormand--Prince (DOPRI) method, and so forth~\cite{DORMAND198019}. Fig.~\ref{fig:archi3} shows the typical architecture of NODEs. There is a feature extraction layer which produces the initial value $\bm{z}(0)$ and the NODE layer evolves it to $\bm{z}(1)$. This architecture is much simpler than that of NCDEs in Fig.~\ref{fig:archi}.

In general, ODE solvers discretize time variable $t$ and convert an integral into many steps of additions. For instance, the explicit Euler method can be written as follows in a step:
\begin{align}\label{eq:euler}
\bm{z}(t + h) = \bm{z}(t) + h \cdot f(\bm{z}(t);\bm{\theta}_f),
\end{align}where $h$, which is usually smaller than 1, is a pre-determined step size of the Euler method.The DOPRI method uses a much more sophisticated method to update $\bm{z}(t + h)$ from $\bm{z}(t)$ and dynamically controls the step size $h$. However, those ODE solvers sometimes incur unexpected numerical instability~\cite{zhuang2020adaptive}. For instance, the DOPRI method sometimes keeps reducing the step size $h$ and eventually, an underflow error is produced. To prevent such unexpected problems, several countermeasures were also proposed.

Instead of the backpropagation method, the adjoint sensitivity method is used to train NODEs for its efficiency and theoretical correctness~\cite{NIPS2018_7892}. After letting $\bm{a}_{\bm{z}}(t) = \frac{d \mathcal{L}}{d \bm{z}(t)}$ for a task-specific loss $\mathcal{L}$, it calculates the gradient of loss w.r.t model parameters with another reverse-mode integral as follows:\begin{align*}\nabla_{\bm{\theta}_f} \mathcal{L} = \frac{d \mathcal{L}}{d \bm{\theta}_f} = -\int_{t_m}^{t_0} \bm{a}_{\bm{z}}(t)^{\mathtt{T}} \frac{\partial f(\bm{z}(t);\bm{\theta}_f)}{\partial \bm{\theta}_f} dt.\end{align*}

$\nabla_{\bm{z}(0)} \mathcal{L}$ can also be calculated in a similar way and we can propagate the gradient backward to layers earlier than the ODE if any. It is worth of mentioning that the space complexity of the adjoint sensitivity method is $\mathcal{O}(1)$ whereas using the backpropagation to train NODEs has a space complexity proportional to the number of DOPRI steps. Their time complexities are similar or the adjoint sensitivity method is slightly more efficient than that of the backpropagation. Therefore, we can train NODEs efficiently.

\subsection{Neural CDEs}\label{sec:ncde}
One limitation of NODEs is that $\bm{z}(t_1)$ is, given $\bm{\theta}_f$, decided by $\bm{z}(t_0)$, which degrades the representation learning capability of NODEs. To this end, NCDEs create another path $X(t)$ from underlying time-series data and $\bm{z}(t_1)$ is decided by both $\bm{z}(t_0)$ and $X(t)$~\cite{DBLP:conf/nips/KidgerMFL20}. The initial value problem (IVP) of NCDEs is written as follows:
\begin{align}
    \bm{z}(t_1) &= \bm{z}(t_0) + \int_{t_0}^{t_1} f(\bm{z}(t);\bm{\theta}_f) dX(t),\\
                &= \bm{z}(t_0) + \int_{t_0}^{t_1} f(\bm{z}(t);\bm{\theta}_f) \frac{dX(t)}{dt} dt,
\end{align}where $X(t)$ is a natural cubic spline path of underlying time-series data. In particular, the integral problem is called \emph{Riemann–Stieltjes integral} (whereas NODEs use the Riemann integral). Differently from NODEs, $f(\bm{z}(t);\bm{\theta}_f)$, which we call \emph{CDE function}, is to approximate $\frac{d \bm{z}(t)}{d X(t)}$. Whereas other methods can be used for $X(t)$, the original authors of NCDEs recommend the natural cubic spline method for its suitable characteristics to be used in NCDEs: i) it is twice differentiable, ii) its computational cost is not much, and iii) $X(t)$ becomes \emph{continuous} w.r.t. $t$ after the interpolation.
 
\subsection{Attention}
Attention, which is to extract key information from data, is one of the most influential concepts in deep learning. It had been widely studied for computer vision and natural language processing~\cite{DBLP:journals/corr/BahdanauCB14,7780872,10.5555/3295222.3295349, spratling2004attention}. After that, it was quickly spread to other fields~\cite{7243334,kim2018attentive,9194070}.

The attention concept can be well explained by the visual perception system of human, which focuses on strategical parts of input while ignoring irrelevant signals~\cite{pmlr-v37-xuc15}. Several different types of attention and their applications have been proposed. A self-attention method was proposed for language understanding~\cite{shen2018disan}. Kiela et al. proposed a general purpose text representation learning method with attention in~\cite{kiela2018dynamic}. For machine translation, a soft attention mechanism was proposed by Bahdanau et al.~\cite{DBLP:journals/corr/BahdanauCB14}. A co-attention method was used for visual question answering~\cite{10.5555/3157096.3157129}. We encourage referring to survey papers~\cite{DBLP:journals/corr/abs-1904-02874,10.1145/3363574} for more detailed descriptions. Like this, there exist many different types and applications of attention.

%To our knowledge, however, it had not been actively studied yet for NCDEs because NCDEs are a relatively new paradigm of designing neural networks and it is not straightforward how to integrate attention into them. 
To our knowledge, attention has not yet been actively studied for NCDEs. This is likely due to the fact that they are a relatively new paradigm of designing neural networks, and that it is not straightforward how to integrate attention into them. On the other hand, several attention-related designs exist for NODEs. ETN-ODE~\cite{gao2020explainable} uses attention in its feature extraction layer before the NODE layer. It does not propose any new NODE model that is \emph{internally} combined with attention. In other words, they use attention to derive $\bm{z}(0)$ in their feature extraction layer, and then use a standard NODE layer to evolve $\bm{z}(0)$. %It is the feature extraction layer which has attention in ETN-ODE. In this regard, it is hard to say that ETN-NODE is an attention-based NODE model.
As the concept of attention is only applied to the extraction layer, it is difficult to classify ETN-ODE as an attention-based NODE model.

As another example, ACE-NODE proposed to augment $\bm{z}(t)$ with another ODE producing attention values. In NODEs, $\bm{z}(t)$ means a hidden vector at time $t$. To help a downstream task with attention, $\bm{z}(t + s)$ should be derived from $\bm{z}(t)$ aided by the attention at time $t$. To this end, its ODE state is extended to $[\bm{z}(t), \bm{z}'(t)]$, where $\bm{z}(t)$ is the main ODE and $\bm{z}'(t)$ is the attention ODE. They evolve both ODEs simultaneously and, while updating $\bm{z}(t)$, it refers to the attention value. They naturally co-evolve rather than evolving independently. As of now, ACE-NODE is the work most similar to our ANCDE. %However, ACE-NODE proposed a method for NODEs whereas we extend NCDEs with attention.
However, the two works differ in that ACE-NODE proposes a novel NODE architecture, while our work extend NCDEs with attention.

\section{Proposed Method}
We propose the method of \underline{\textbf{A}}ttentive \underline{\textbf{N}}eural \underline{\textbf{C}}ontrolled \underline{\textbf{D}}ifferential \underline{\textbf{E}}quation (ANCDE), where we adopt two NCDEs: the bottom NCDE for calculating attention values and the top NCDE for time-series classification/forecasting.

\subsection{Overall Workflow}

\begin{figure*}
    \centering
    \includegraphics[width=0.8\textwidth]{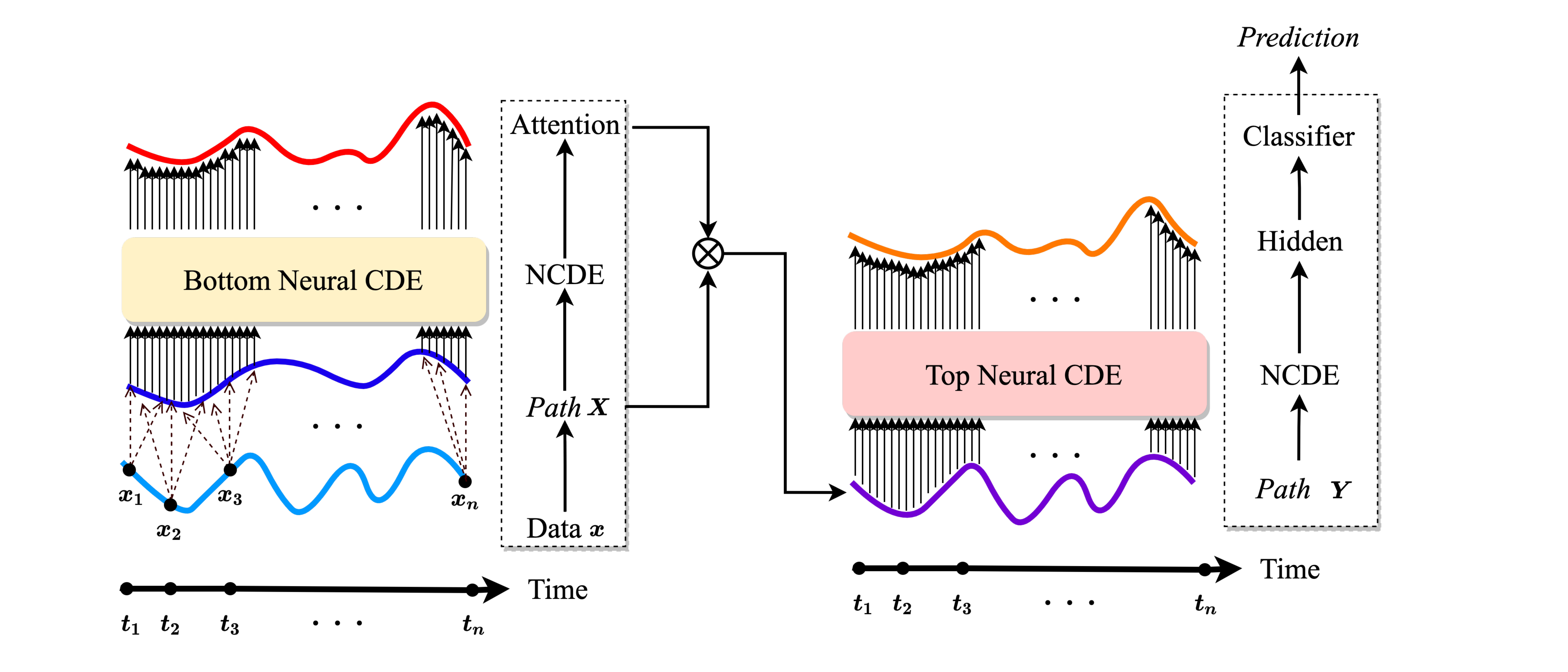}
    \caption{The detailed architecture of our proposed attentive NCDE (ANCDE). From the raw time-series data $x(t)$, the continuous path $X(t)$ is created by the natural cubic spline algorithm. The bottom NCDE reads $X(t)$ to produce attention values. The path $Y(t)$ is defined as the (element-wise) multiplication of $X(t)$ and the attention value at time $t$. The top NCDE produces the last hidden vector with $Y(t)$. Our framework supports both classification and regression (although we assume classification in this figure).}
    \label{fig:archi2}
\end{figure*}

Fig.~\ref{fig:archi2} shows the detailed architecture of our proposed ANCDE. Given a time-series data $\{(x_0, t_0), (x_1, t_1), \cdots\}$ annotated with observation time, we first construct a continuous path $X(t)$ with the natural cubic spline algorithm. We then feed $X(t)$ into the bottom NCDE which produces attention values for each time-point $t$. Another path $Y(t)$ is created by the (element-wise) multiplication of the attention and $X(t)$ in Eq.~\eqref{eq:y} and fed into the top NCDE which produces the last hidden vector. There is one more classification layer in Eq.~\eqref{eq:cla} which produces the final prediction.

Even though each raw data point $(x_i, t_i)$ is observed in a discrete and irregular fashion, the path $X(t)$ is continuous and we note that $X(t_i) = x_i$, where $t_t$ is the time-point when $x_i$ is observed. For other non-observed time-points, the natural cubic spline algorithm interpolates nearby observed data.

\medskip\noindent{\textbf{NCDE vs. NODE: }} The key difference between NCDEs and NODEs lies in the existence of the path $X(t)$. When calculating $X(t)$ in NCDEs, in particular, the natural cubic spline considers some selected future observations $\{x_{t'}\}$, where $t' \geq t$, in conjunction with its current and past observations, which is not the case in NODEs. Therefore, NCDEs are considered as a more general method than NODEs~\cite{NEURIPS2020_4a5876b4}. In particular, NCDEs reduce to NODEs when $X(t) = t$, i.e., the identity function w.r.t. time.

\subsection{Attentive NCDE}
\subsubsection{Bottom NCDE for attention values} The bottom NCDE can be written as follows:
\begin{align}
    \bm{h}(t_1) &= \bm{h}(t_0) + \int_{t_0}^{t_1} f(\bm{h}(t);\bm{\theta}_f) dX(t),\\
                &= \bm{h}(t_0) + \int_{t_0}^{t_1} f(\bm{h}(t);\bm{\theta}_f) \frac{dX(t)}{dt} dt,
\end{align}where $\bm{h}(t)$ means a hidden vector of attention. We later derive attention at time $t$ from $\bm{h}(t)$. We note that this bottom NCDE is equivalent to the original NCDE setting.

We support two different attention concepts: i) time-wise attention $a(t) \in \mathbb{R}$ and ii) element-wise attention $\bm{a}(t) \in \mathbb{R}^{\dim(X(t))}$. In the former type, the output size of the fully-connected layer $\mathtt{FC}_1$ is 1 and $a(t) = \sigma(\mathtt{FC}_1(\bm{h}(t)))$ is a scalar value. In the latter type, $\bm{a}(t) = \sigma(\bm{h}(t))$ is a vector.

We also support three different types of $\sigma$: i) the soft attention with the original sigmoid, ii) the hard attention with the original sigmoid followed by the rounding function, iii) the straight-through estimator, i.e., the hard attention with the sigmoid slope annealing. We omit the description of the soft attention since we use the original sigmoid. However, the hard attention has different forward and backward path definitions as follows:
\begin{align}
    \sigma(x) = round(sigmoid(x)),\textrm{ for the forward path},\\
    \nabla \sigma(x) = \nabla sigmoid(x),\textrm{ for the backward path}.
\end{align}

With the straight-through estimator, we have
\begin{align}
    \sigma(x) = round(sigmoid(\tau x)),\textrm{ for the forward path},\\
    \nabla \sigma(x) = \nabla sigmoid(\tau x),\textrm{ for the backward path},
\end{align}where the temperature $\tau \geq 1.0$ is a scalar multiplicative term to control the slope of the sigmoid function. We note that, with a large enough $\tau$, the sigmoid slope becomes close to that of the rounding function. Therefore, we keep increasing $\tau$ slowly as training goes on. We add 0.12 to $\tau$ every epoch after initializing it to 1 at the beginning.

In total, therefore, we can define six attention models, i.e., six combinations from the two attention types and the three different definitions of $\sigma$, as shown in Table~\ref{tbl:attn}.

\begin{table}[t]
\centering
\setlength{\tabcolsep}{2pt}
\caption{Six different types of attention in our framework}\label{tbl:attn}
\begin{tabular}{ccc}
\hline
Type & Time/Element-wise & $\sigma$ \\ 
\hline
SOFT-TIME & Time-wise (i.e., $a(t) \in \mathbb{R}$) & Sigmoid \\ 
HARD-TIME & Time-wise (i.e., $a(t) \in \mathbb{R}$) & Hard \\
STE-TIME & Time-wise (i.e., $a(t) \in \mathbb{R}$) & Straight-through \\
SOFT-ELEM & Element-wise (i.e., $\bm{a}(t) \in \mathbb{R}^{\dim(X(t))}$) & Sigmoid \\ 
HARD-ELEM & Element-wise (i.e., $\bm{a}(t) \in \mathbb{R}^{\dim(X(t))}$) & Hard \\
STE-ELEM & Element-wise (i.e., $\bm{a}(t) \in \mathbb{R}^{\dim(X(t))}$) & Straight-through \\
\hline
\end{tabular}
\end{table}

\subsubsection{Top NCDE for time-series classification/forecasting} The top NCDE can be written as follows:
\begin{align}
    \bm{z}(t_1) &= \bm{z}(t_0) + \int_{t_0}^{t_1} g(\bm{z}(t);\bm{\theta}_g) dY(t),\\
                &= \bm{z}(t_0) + \int_{t_0}^{t_1} g(\bm{z}(t);\bm{\theta}_g) \frac{dY(t)}{dt} dt,\label{eq:top}\\
    Y(t)   &= \begin{cases} a(t)X(t),\textrm{ if time-wise attention}\\ \label{eq:y}
    \bm{a}(t)\otimes X(t),\textrm{ if element-wise attention}
    \end{cases},
\end{align}where $\otimes$ stands for the element-wise multiplication, and $Y(t)$ is the (element-wise) multiplication of the attention and $X(t)$. Therefore, $Y(t)$ contains information selected by the bottom NCDE. In this regard, the top NCDE is able to focus on useful information only and the performance of the downstream machine learning task can be improved.

%  (we assume the time-wise attention for this description for simplicity but without loss of generality. For the element-wise attention, the following equation can be readily modified accordingly.)
 
We further derive Eq.~\eqref{eq:top} into a more tractable form as follows:
\begin{align} \label{eq:top_same}
    \bm{z}(t_1) = \bm{z}(t_0) + &\int_{t_0}^{t_1} g(\bm{z}(t);\bm{\theta}_g) \frac{dY(t)}{dt} dt,
    % = \bm{z}(t_0) + &\int_{t_0}^{t_1} g(\bm{z}(t);\bm{\theta}_g) \Big(\frac{dX(t)}{dt}a(t) +\\
    % & X(t)a(t)(1-a(t))\frac{dh(t)}{dt}\Big) dt,
\end{align} where for the time-wise attention,
\begin{align}\label{eq:yt_timewise}
    \frac{dY(t)}{dt} = \frac{dX(t)}{dt}a(t) + X(t)a(t)(1-a(t))\frac{d\mathtt{FC}_1}{d\bm{h}(t)}\frac{d\bm{h}(t)}{dt},
\end{align} or for the element-wise attention,
\begin{align}\label{eq:yt_elementwise}
    \frac{dY(t)}{dt} = \frac{dX(t)}{dt}\bm{a}(t) + X(t)\bm{a}(t)(1-\bm{a}(t))\frac{d\bm{h}(t)}{dt}.
\end{align}

% \begin{align} \label{eq:yt}
% \frac{dY(t)}{dt}   &= \begin{cases} \frac{dX(t)}{dt}a(t) + X(t)a(t)(1-a(t))\frac{d\mathtt{FC}_1}{dh(t)}\frac{dh(t)}{dt},\\ \textrm{ if time-wise attention}\\ \label{eq:y}
%     \frac{dX(t)}{dt}a(t) + X(t)a(t)(1-a(t))\frac{dh(t)}{dt},\\ \textrm{ if element-wise attention} 
%     \end{cases},
% \end{align}

We note that Eqs.~\eqref{eq:yt_timewise},~\eqref{eq:yt_elementwise} are from the derivative of the sigmoid function. In these derivations, we assume the soft attention. However, this does not theoretically break the working mechanism of the hard attention and the staraight-through estimator. The hard attention generates values in \{0,1\} whereas values in $[0,1]$ are produced by the soft attention. Therefore, the hard attention is still in the valid range. For instance, suppose the time-wise attention. When $a(t) = 0$ (resp. $a(t)=1$) by the hard attention, $\frac{dY(t)}{dt} = 0$ (resp. $\frac{dY(t)}{dt} = \frac{dX(t)}{dt}$), which exactly corresponds to our attention motivation, i.e., the top NCDE focuses on the values selected by the bottom NCDE. The straight-through estimator is considered as a variant of the hard attention with the temperature annealing and it also produces values in $\{0,1\}$. Therefore, there are no problems in Eqs.~\eqref{eq:yt_timewise},~\eqref{eq:yt_elementwise} to be used for all the three attention types.

\subsection{Training Algorithm}
We use the adjoint backpropagation method \cite{NIPS2018_7892, pontryagin1962the, giles2000an, hager2001runge} to train NCDEs, which requires a memory of $\mathcal{O}(L+H)$ where $L = t_1 - t_0$, i.e., the integral time space, and $H$ is the size of the vector field defined by an NCDE. Therefore, the training at Lines~\ref{alg:train1} to~\ref{alg:train3} of Alg.~\ref{alg:train} is done with the adjoint method. However, our framework has two NCDEs and as a result, it needs a memory of $\mathcal{O}(2L + H_f + H_g)$ where $H_f$ and $H_g$ are the sizes of the vector fields created by the bottom and the top NCDEs, respectively. 

\begin{algorithm}[t]
\SetAlgoLined
\caption{How to train ANCDE}\label{alg:train}
\KwIn{Training data $D_{train}$, Validating data $D_{val}$, Maximum iteration number $max\_iter$}
Initialize $\bm{\theta}_f$, $\bm{\theta}_g$, and other parameters $\bm{\theta}_{others}$ if any, e.g., the parameters of the feature extractor and the classifier;

$k \gets 0$;

\While {$k < max\_iter$}{
    Train $\bm{\theta}_{others}$ after fixing $\bm{\theta}_f$ and $\bm{\theta}_g$\;\label{alg:train1}
    
    Train $\bm{\theta}_f$ after fixing $\bm{\theta}_{others}$ and $\bm{\theta}_g$\;\label{alg:train2}
    
    Train $\bm{\theta}_g$ after fixing $\bm{\theta}_f$ and $\bm{\theta}_{others}$\;\label{alg:train3}
    
    Validate and update the best parameters, $\bm{\theta}^*_f$, $\bm{\theta}^*_g$, and $\bm{\theta}^*_{others}$, with $D_{val}$\;
    
    $k \gets k + 1$;
}
\Return $\bm{\theta}^*_f$, $\bm{\theta}^*_g$, and $\bm{\theta}^*_{others}$;
\end{algorithm}

\medskip\noindent{\textbf{Well-posedness of the Problem:}} The well-posedness\footnote{A well-posed problem means i) its solution uniquely exists, and ii) its solution continuously changes as input data changes.} of NCDEs, given a fixed path, was already proved in \cite[Theorem 1.3]{lyons2004differential} under the mild condition of the Lipschitz continuity. We show that our ANCDEs are also well-posed problems. Almost all activations, such as ReLU, Leaky ReLU, SoftPlus, Tanh, Sigmoid, ArcTan, and Softsign, have a Lipschitz constant of 1. Other common neural network layers, such as dropout, batch normalization and other pooling methods, have explicit Lipschitz constant values. Therefore, the Lipschitz continuity of $f$ and $g$ can be fulfilled in our case. Therefore, it is a well-posed training problem for the bottom NCDE producing attention values given other fixed parts (Line~\ref{alg:train2}). Given a fixed path $Y(t)$, then it becomes well-posed again for the top NCDE (Line~\ref{alg:train3}). Therefore, our training algorithm solves a well-posed problem so its training process is stable in practice.

In our experiments, we solve time-series classification problems and use the standard cross-entropy loss with the final hidden vector $\bm{z}(t_1)$, followed by a classification layer as follows:
\begin{align}\label{eq:cla}
    \hat{y} = \phi(\mathtt{FC}_2(\bm{z}(t_1))),
\end{align}where $\hat{y}$ is a predicted class label and $\phi$ is softmax activation. The output size of $\mathtt{FC}_2$ is the same as the number of classes and we use the standard cross-entropy loss. For regression, the output size of $\mathtt{FC}_2$ is the same as the size of target prediction and % we do not use the softmax activation. We use the mean squared error loss.
we utilize mean squared error loss.

\begin{table}[t]
\centering
\caption{The architecture of the CDE function $f$ in the bottom NCDE for Character Trajectories. \texttt{FC} stands for the fully-connected layer, $\rho$ means the rectified linear unit (ReLU), and $\xi$ is the hyperbolic tangent (tanh).}\label{tbl:f1}
\begin{tabular}{cccc}
\hline
Layer & Design & Input Size & Output Size \\ 
\hline
1 & \texttt{FC}& $32 \times $4  & $32 \times $10  \\ 
2 & $\rho$(\texttt{FC}) & $32 \times $10  & $32 \times $20 \\ 
3 & $\rho$(\texttt{FC}) & $32 \times $20  & $32 \times $20 \\ 
4 & $\rho$(\texttt{FC}) & $32 \times $20  & $32 \times $20 \\ 
5 & $\xi$(\texttt{FC}) & $32 \times $20  & $32 \times $16 \\ 
\hline
\end{tabular}
\end{table}

\begin{table}[t]
\centering
\caption{The architecture of the CDE function $g$ in the top NCDE for Character Trajectories}\label{tbl:g1}
\begin{tabular}{cccc}
\hline
Layer & Design & Input Size & Output Size \\ 
\hline
1 & \texttt{FC}& $32 \times $40  & $32 \times $40  \\ 
2 & $\rho$(\texttt{FC}) & $32 \times $40  & $32 \times $40 \\ 
3 & $\rho$(\texttt{FC}) & $32 \times $40  & $32 \times $40 \\ 
4 & $\xi$(\texttt{FC}) & $32 \times $40  & $32 \times $160 \\ 
 
\hline
\end{tabular}
\end{table}

\begin{table}[t]
\centering
\caption{The architecture of the CDE function $f$ in the Bottom NCDE for PhysioNet sepsis}\label{tbl:f2}
\begin{tabular}{cccc}
\hline
Layer & Design & Input Size & Output Size \\ 
\hline
1 & \texttt{FC}& $1024 \times $69  & $1024 \times $20  \\ 
2 & $\rho$(\texttt{FC}) & $1024 \times $20  & $1024 \times $20 \\ 
3 & $\rho$(\texttt{FC}) & $1024 \times $20  & $1024 \times $20 \\ 
4 & $\rho$(\texttt{FC}) & $1024 \times $20  & $1024 \times $20 \\
5 & $\xi$(\texttt{FC}) & $1024 \times $20 & $1024 \times $4761\\
 
\hline
\end{tabular}
\end{table}

\begin{table}[t]
\centering
\caption{The architecture of the CDE function $g$ in the Top NCDE for PhysioNet sepsis}\label{tbl:g2}
\begin{tabular}{cccc}
\hline
Layer & Design & Input Size & Output Size \\ 
\hline
1 & \texttt{FC}& $1024 \times $49  & $1024 \times $49  \\ 
2 & $\rho$(\texttt{FC}) & $1024 \times $49  & $1024 \times $49 \\ 
3 & $\rho$(\texttt{FC}) & $1024 \times $49  & $1024 \times $49 \\ 
4 & $\rho$(\texttt{FC}) & $1024 \times $49  & $1024 \times $49 \\
5 & $\xi$(\texttt{FC}) & $1024 \times $49 & $1024 \times $3381\\

\hline
\end{tabular}
\end{table}

\begin{table}[t]
\centering
\caption{The architecture of the CDE function $f$ in the Bottom NCDE for Google Stock}\label{tbl:f3}
\begin{tabular}{cccc}
\hline
Layer & Design & Input Size & Output Size \\ 
\hline
1 & \texttt{FC}& $200 \times $7  & $200 \times $8  \\ 
2 & $\rho$(\texttt{FC}) & $200 \times $8  & $200 \times $4 \\ 
3 & $\rho$(\texttt{FC}) & $200 \times $4  & $200 \times $4 \\ 
4 & $\rho$(\texttt{FC}) & $200 \times $4  & $200 \times $4 \\
5 & $\xi$(\texttt{FC}) & $200 \times $4 & $200 \times $49\\
 
\hline
\end{tabular}
\end{table}
\begin{table}[t]
\centering
\caption{The architecture of the CDE function $g$ in the Top NCDE for Google Stock}\label{tbl:g3}
\begin{tabular}{cccc}
\hline
Layer & Design & Input Size & Output Size \\ 
\hline
1 & \texttt{FC}& $200 \times $32  & $200 \times $32  \\ 
2 & $\rho$(\texttt{FC}) & $200 \times $32  & $200 \times $32 \\ 
3 & $\rho$(\texttt{FC}) & $200 \times $32  & $200 \times $32 \\ 
4 & $\xi$(\texttt{FC}) & $200 \times $32  & $200 \times $224 \\
\hline
\end{tabular}
\end{table}

\section{Experiments}
In this section, we describe our experimental environments and results. We conduct experiments on image classification and time-series forecasting. All experiments were conducted in the following software and hardware environments: \textsc{Ubuntu} 18.04 LTS, \textsc{Python} 3.6.6, \textsc{Numpy} 1.18.5, \textsc{Scipy} 1.5, \textsc{Matplotlib} 3.3.1, \textsc{PyTorch} 1.2.0, \textsc{CUDA} 10.0, \textsc{NVIDIA} Driver 417.22, i9 CPU, and \textsc{NVIDIA RTX Titan}. We repeat the training and testing procedures with five different random seeds and report their mean and standard deviation accuracy. Our source code and data can be found at {\color{blue}\url{https://github.com/sheoyon-jhin/ANCDE}}.

\subsection{Datasets}
\medskip\noindent{\textbf{Character Trajectories: } The Character Trajectories dataset from the UEA time-series classification archive~\cite{bagnall2018uea} %is one of our time-series classification experiments. This dataset contains information about the X and Y-axes positions and the pen tip force of the Latin alphabets written on a tablet with a sampling frequency of 200Hz. This dataset consists of 2,858 character samples. 
contains 2,858 pen tip trajectory samples of Latin alphabets written on a tablet. They contain information about the X and Y-axes position and pen tip force with a sampling frequency of 200Hz.
All data samples were truncated to a time-series length of 182, the shortest length in the dataset. Each sample is represented by a time-series of 3 dimensional vectors, i.e., X and Y-axes positions and pen tip force. The goal is to classify each sample into one of the 20 different class labels (`a', `b', `c', `d', `e', `g', `h', `l', `m', `n', `o', `p', `q', `r', `s', `u', `v', `w', `y', `z' ). Note that some alphabet characters are missing in this dataset.}

\medskip\noindent{\textbf{PhysioNet Sepsis: } The PhysioNet 2019 challenge on sepsis prediction~\cite{9005736,article } is also used for our time-series classification experiments. %Sepsis refers to a body-wide phenomenon caused by bacteria or bacterial toxins in the blood and is life-threatening. In the U.S., about 1.7 million people developed sepsis and 270,000 of them died at 2018. More than a third of people who die in U.S. hospitals have sepsis. Early predictions of sepsis could potentially save lives, so this experiment is meaningful. The dataset consists of a total of 40,335 cases and was measured for patients staying in intensive care units.
Sepsis is a life-threatening condition caused by bacteria or bacterial toxins in the blood. In the U.S., about 1.7 million people develop sepsis and 270,000 die from sepsis in a year. More than a third of patients who die in U.S. hospitals have sepsis. Early prediction of sepsis could potentially save lives, so this experiment is especially meaningful. The dataset consists of 40,335 cases of patients staying in intensive care units.
There are 34 time-dependent variables, such as heart rate, oxygen saturation, body temperature, and so on. The last variable is the class label, which indicates the onset of sepsis according to the sepsis-3 definition, where ``1" indicates sepsis and ``0" means no sepsis, and ``nan'' means no information. The goal is to classify whether each patient has sepsis or not after removing the ``nan" class.}

% \paragraph{PEMS-SF} The PEMS-SF dataset~\cite{Dua:2019} is about the occupancy of different highway lanes in the San Francisco Bay Area. This dataset measured daily lane occupancy rates on highways, consisting of 440 time-series samples. Each sample, which represents daily lane occupancy rates, has a length of 144 time-stamped observations, and each observation (value) contains 963 sensing values, i.e., a 963 dimensional vector. Each sample has a ground-truth label of 1 through 7, representing Monday to Sunday. The goal is to classify on what day of the week the data for that instance was written.

\medskip\noindent{\textbf{Google Stock: } The Google Stock dataset includes trading volumes in conjunction with the high, low, open, close, and adjusted closing prices from 2004 to 2019~\cite{NEURIPS2019_c9efe5f2}. This dataset is used for time-series forecasting experiments. CDEs were originally developed in the field of mathematical finance to predict various financial time-series values~\cite{Lyons1998}. Therefore, we expect that our ANCDE shows the best appropriateness for this task. The goal is to predict, given past several days of time-series values, the high, low, open, close, and adjusted closing prices at the very next day.}

\subsection{Experimental Methods}
We conduct classification and forecasting (regression) experiments to verify the efficacy of the proposed attention-based method. Our experiments can be classified into the following two types: i) regular time-series experiments, and ii) irregular time-series experiments. For the regular time-series experiments, we use all the data provided in each dataset without dropping any observations. For the irregular time-series experiments, however, we randomly drop 30\%, 50\%, and 70\% of observations from each individual time-series sample and then try to predict, e.g., we read the stock price observations in past random 70\% days for Google and predict next prices. 

%The only exception is the PhysioNet Sepsis dataset, where random 90\% of observations were already removed before being released by the authority. By removing many observations and individual information from the dataset, they successfully protected the privacy of the people who had contributed to this dataset. We perform only irregular time-series classification with this dataset.
The only exception is the PhysioNet Sepsis datatset, where 90\% of the observations were randomly removed by the data collectors in order to protect the privacy of the people who had contributed to the dataset. Hence, we only perform irregular time-series classification with this data.

Different performance metrics are utilized for each task. For classification with balanced classes we use accuracy, whereas AUCROC is utilized in the case of imbalanced classification. For forecasting problems, we consider mean absolute error (MAE) or mean squared error (MSE).

\subsection{Baselines}
For regular time-series forecasting and classification experiments, we compare our method with the following methods:
\begin{enumerate}
    \item RNN, LSTM~\cite{sepp1997long}, GRU~\cite{chung2014empirical} are all recurrent neural network-based models which are able to handle sequential data. LSTM is designed to learn long-term dependencies and GRU uses gating mechanisms to control the flow of information.
    \item GRU-ODE~\cite{debrouwer2019gruodebayes,jordan2019gated} is a NODE analogous to GRU. This model is a continuous counterpart to GRU.
    \item ODE-RNN~\cite{rubanova2019latent} is an extension of NODE by adopting a GRU between observations. This combination of NODE and GRU is known as an \emph{jump}~\cite{herrera2021neural}.
    \item Latent-ODE~\cite{NIPS2018_7892} is a time-series model in which the latent state follows a NODE. The work in \cite{rubanova2019latent} uses the ODE-RNN to replace the recognition network of the original Latent-ODE model. We also use the ODE-RNN as a recognition network.
    \item ACE-NODE~\cite{jhin2021acenode} is the state-of-the-art attention-based NODE model which has dual co-evolving NODEs.
\end{enumerate}

For the irregular time-series forecasting and classification experiments, we compare our method with the following methods:
\begin{enumerate}
    \item GRU-$\Delta t$~\cite{NEURIPS2020_4a5876b4} is the GRU with the time difference between observations additionally used as an input.
    \item GRU-D~\cite{che2016recurrent} is a modified version of GRU-$\Delta t$ with a learnable exponential decay between observations.
    \item GRU-ODE, ODE-RNN, Latent-ODE, and ACE-NODE can also be utilized in irregular time-series experiments as they were developed with consideration of irregular time-series patterns.
    %GRU-ODE, ODE-RNN, Latent-ODE, and ACE-NODE can also be used for the irregular time-series experiments since those models are designed to be able to process irregular time-series patterns.
\end{enumerate}

\begin{table}[t]
\centering
\caption{Test Accuracy (mean ± std, computed across five runs) on Regular Character Trajectories}\label{tbl:charactertrajectory0}
\begin{tabular}{ccc}
\hline
Model & Test Accuracy &Memory Usage(MB) \\ 
\hline
RNN & 0.162 ± 0.038 & 0.78\\
LSTM & 0.499 ± 0.089 &0.68\\
GRU & 0.909 ± 0.086 &0.77\\
GRU-ODE & 0.778 ± 0.091& 1.50\\
ODE-RNN & 0.427 ± 0.078& 14.8\\
Latent-ODE & 0.954 ± 0.003& 181\\
ACE-NODE  & 0.981 ± 0.001 & 113\\
NCDE & 0.974 ± 0.004&1.38\\
\hline
\textbf{ANCDE} & \textbf{0.991 ± 0.002}&2.02\\
\hline
\end{tabular}
\end{table}

\subsection{Hyperparameters}

We test the following common hyperparameters for our method and other baselines:
\begin{enumerate}
    \item In Character Trajectories, we train for 200 epochs with a batch size of 32, a learning rate of $\{\num{1.0e-4}, \num{5.0e-4}, \num{1.0e-3}, \num{5.0e-3}\}$ and a hidden vector dimension of $\{10, 20, 40, 50, 60\}$ and set the ODE function to have $\{3, 4, 5\}$ layers;
    
    % In GRU-ODE, GRU-$\Delta t$, GRU-D  baselines, we used 47 hidden vector dimension and batch size of 32. The ODE function of various NODE-based baselines have 1 fully connected (FC) layers. In Latent ODE(ODE Enc.) and ACE-NODE we used 15 latent dimensions in the generative model, 40 dimensions in recognition model and batch size of 32. The ODE function of various NODE-based baselines have 3 fully connected (FC) layers. In ODE-RNN and Neural CDE and Attentive Neural CDE(our method) we used 40 hidden vector dimension and 3 fully connected (FC) layers.
    
    \item In PhysioNet Sepsis, we train for 150 epochs with a batch size 1024, a learning rate of $\{\num{1.0e-5},\num{5.0e-5},\num{1.0e-4}, \num{5.0e-4}\}$, a hidden vector dimension of $\{30,40,50,60\}$, and set the ODE function to have $\{2,3,4,5\}$ layers;
    
    \item In Google Stock, we train for 150 epochs with a batch size 200, a learning rate of $\{\num{1.0e-2},\num{1.0e-3},\num{1.0e-4}\}$, a hidden vector dimension of $\{30,40,50,60\}$, and set the ODE function to have $\{2,3,4\}$ layers;
\end{enumerate}

For reproducibility, we report the best hyperparameters for our method as follows:
\begin{enumerate}
    \item In Character Trajectories, we set the learning rate to \num{1.0e-3};
    \item In PhysioNet Sepsis, we set the learning rate to \num{1.0e-5};
    \item In Google Stock, we set the learning rate to \num{1.0e-3}.
    \item In Tables~\ref{tbl:f1} to~\ref{tbl:g3}, we also clarify the network architecture of the CDE functions, $f$ and $g$, i.e., the main networks modeling  $\frac{d \bm{h}(t)}{d X(t)}$ and $\frac{d \bm{z}(t)}{d Y(t)}$, respectively. Note that these tables denote the number of layers and their input/output dimensions. We specifically use these settings because they make the size of our model comparable to other baselines, resulting in comparable GPU memory usage during inference.
\end{enumerate}

\subsection{Experimental Results}
\medskip\noindent{\textbf{Character Trajectories (Regular): } We first introduce our experimental results with regular time-series classification on Character Trajectories. Table~\ref{tbl:charactertrajectory0} summarizes the results. All RNN-based methods except GRU show unreliable accuracy. In particular, RNN shows unreasonable accuracy in comparison with others due to its low representation learning capability. However, GRU shows good accuracy. In general, NODE and NCDE-based methods show good accuracy. ACE-NODE and NCDE show almost the same mean accuracy. The attention mechanism of ACE-NODE significantly increases the accuracy in comparison with other NODE-based models. %However, our ANCDE marks the clearly best accuracy with a non-trivial margin. It shows a near-perfect accuracy of 99.2\%.
ANCDE greatly outperforms other methods by a wide margin, showing a near-perfect accuracy of 99.1\%.
}
% UEA

\begin{table}[t]
% \footnotesize
\setlength{\tabcolsep}{2pt}
\centering
\caption{Test Accuracy on Irregular Character Trajectories}\label{tbl:charactertrajectory1}
{
% {\scriptsize
\begin{tabular}{ccccc} \hline
\multirow{2}{*}{Model} & \multicolumn{3}{c}{Test Accuracy} & \multirow{2}{*}{\begin{tabular}[c]{@{}c@{}}Memory\\Usage (MB)\end{tabular}} \\ \cline{2-4}& 30\% dropped & 50\% dropped & 70\% dropped &                                                                              \\  \hline
GRU-ODE & 0.926 ± 0.016 & 0.867 ± 0.039 & 0.899 ± 0.037 & 1.50    
\\
GRU-$\Delta t$ & 0.936 ± 0.020 & 0.913 ± 0.021 & 0.904 ± 0.008 & 15.8    
\\
GRU-D & 0.942 ± 0.021 & 0.902 ± 0.048 & 0.919 ± 0.017 & 17.0    
\\
ODE-RNN & 0.954 ± 0.006 & 0.960 ± 0.003 & 0.953 ± 0.006 & 14.8    
\\
Latent-ODE & 0.875 ± 0.027 & 0.869 ± 0.021 & 0.887 ± 0.590 & 181  
\\
ACE-NODE & 0.876 ± 0.055 & 0.886 ± 0.025 & 0.910 ± 0.032 & 113   
\\
NCDE  & 0.987 ± 0.008 & 0.988 ± 0.002 & 0.986 ± 0.004 & 1.38                                                                          \\ \hline
\textbf{ANCDE} & \textbf{0.992 ± 0.003} & \textbf{0.989 ± 0.001} & \textbf{0.988 ± 0.002} & 2.02                                                            \\ \hline
\end{tabular}%
}

\end{table}

\begin{table}[t]
\setlength{\tabcolsep}{2pt}
\centering
\caption{Test AUCROC on PhysioNet Sepsis}\label{tbl:PhysioNet}
\begin{tabular}{cccccc} \hline
\multicolumn{1}{c}{\multirow{2}{*}{Model}} & \multicolumn{2}{c}{Test AUCROC} & & \multicolumn{2}{c}{Memory Usage (MB)} \\ \cline{2-3}\cline{5-6}
\multicolumn{1}{c}{}  & OI  & No OI  & & OI & No OI \\ \hline
GRU-ODE & 0.852 ± 0.010 & 0.771 ± 0.024 & & 454 & 273  \\
GRU-$\Delta t$ & 0.878 ± 0.006 & 0.840 ± 0.007 & & 837 & 826  \\
GRU-D & 0.871 ± 0.022 & \textbf{0.850 ± 0.013} & & 889 & 878  \\
ODE-RNN & 0.874 ± 0.016 & 0.833 ± 0.020 & & 696 & 686  \\
Latent-ODE & 0.787 ± 0.011 & 0.495 ± 0.002 & & 133 & 126  \\
ACE-NODE & 0.804 ± 0.010 & 0.514 ± 0.003 & & 194 & 218  \\
NCDE  & 0.880 ± 0.006 & 0.776 ± 0.009 & & 244 & 122 \\\hline
\textbf{ANCDE} & \textbf{0.900 ± 0.002} & 0.823 ± 0.003 & & 285 & 129   \\ \hline
\end{tabular}
\end{table}

\begin{table}[t]

\centering

\caption{Test MSE on Regular Google Stock}\label{tbl:google0}
\begin{tabular}{ccc}
\hline
Model & Test MSE &Memory Usage(MB)\\ 
\hline
RNN & 0.002 ± 0.001 &2.95\\
LSTM & 0.004 ± 0.001 &8.19\\
GRU & 0.003 ± 0.000 &8.62\\
GRU-ODE & 0.049 ± 0.001&4.04 \\
GRU-D & 0.039 ± 0.001 &10.2\\
ODE-RNN & 0.029 ± 0.001 &10.9\\
Latent-ODE & 0.002 ± 0.001&4.40 \\
ACE-NODE  & 0.002 ± 0.000 &15.4\\
NCDE & 0.304 ± 0.023&3.89\\
\hline
\textbf{ANCDE} & \textbf{0.001 ± 0.001}&4.49\\
\hline

\end{tabular}
\end{table}

\begin{table}[!t]
\setlength{\tabcolsep}{2pt}
\centering
\caption{Test MSE on Irregular Google Stock}
\label{tbl:google1}

% {\scriptsize
\begin{tabular}{ccccc} \hline
\multirow{2}{*}{Model} & \multicolumn{3}{c}{Test MSE} & \multirow{2}{*}{\begin{tabular}[c]{@{}c@{}}Memory\\Usage (MB)\end{tabular}} \\ \cline{2-4}& 30\% dropped & 50\% dropped & 70\% dropped &                                                                              \\  \hline
GRU-ODE &  0.049 ± 0.001  & 0.054 ± 0.001 & 0.054 ± 0.001 & 4.04  
\\
GRU-$\Delta t$ & 0.039 ± 0.000 & 0.041 ± 0.000 & 0.041 ± 0.000 &  9.58   
\\
GRU-D &0.039 ± 0.001 & 0.041 ± 0.001 & 0.041 ± 0.001 & 10.2
\\
ODE-RNN & 0.029 ± 0.001 & 0.031 ± 0.001 & 0.031 ± 0.000 & 10.9    
\\
Latent-ODE & 0.031 ± 0.002 &0.022 ± 0.009 & 0.022 ± 0.003 & 4.40   
\\
ACE-NODE & 0.029 ± 0.007 & 0.014 ± 0.004 & 0.013 ± 0.001 & 15.4    
\\
NCDE  & 0.304 ± 0.023 & 0.341 ± 0.090 & 0.535 ± 0.584 & 3.89                                                                          \\ \hline
\textbf{ANCDE} & \textbf{0.001 ± 0.002} & \textbf{0.001 ± 0.001} & \textbf{0.001 ± 0.002} & 4.49                                                            \\ \hline
\end{tabular}%

\end{table}

\medskip\noindent{\textbf{Character Trajectories (Irregular): } We also conduct irregular time-series classification experiments on Character Trajectories. For this, we randomly drop 30/50/70\% of values in a time-series sample and classify. Recall that in this dataset, each time-series sample includes a sequence of (X-position, Y-position, Pen tip force) and its class is an alphabet character. Therefore, our irregular time-series classification corresponds to classifying with partial observations on drawing patterns in a tablet.

Table~\ref{tbl:charactertrajectory1} summarizes the results. One interesting point is that for some models, irregular time-series classification accuracy is higher than that of regular time-series classification, e.g., GRU-ODE shows an accuracy of 0.926 for irregular classification tasks vs. 0.778 for regular classification. This is because the classification task does not need fine-grained time-series input. For our ANCDE, in addition, the accuracy after dropping 30\% is the same as that of the regular time-series experiment, which corroborates the fact that relatively coarse-grained observations are enough for this task. Our method consistently shows the best accuracy. Another upside of our method is that its memory requirements are small --- only 2MB of GPU memory is needed for inference.
}

\medskip\noindent{\textbf{PhysioNet Sepsis: } The PhysioNet dataset is basically an irregular time-series dataset since its creator sampled only 10\% of values (with their time stamps) for each patient. Therefore, we do not conduct regular time-series experiments on this dataset. Instead, we consider two different irregular time-series classification cases: i) one with observation intensity (OI), and ii) the other without observation intensity (No OI). When we consider observation intensity, we add an index number for each time-series value. Patients are typically frequently checked when their conditions are bad. Therefore, the intensity, i.e., the number of observations, somehow implies the mortality class label.

Table~\ref{tbl:PhysioNet} contains the experimental results. With OI, many methods show good AUCROC scores. All GRU-based irregular time-series classification models show scores of 0.852 to 0.878. However, NODE-based methods outperform them. In general, NCDE-based methods are better than NODE-based models. Our ANCDE shows the best AUCROC score with a little memory overhead in comparison with NCDE.
}

\begin{table}[t]
\setlength{\tabcolsep}{2pt}
\centering
\caption{Ablation Study on Irregular Character Trajectories}\label{tbl:abl_charactertrajectory2}
\begin{tabular}{cccc}
\hline
\multirow{2}{*}{Model}   & \multicolumn{3}{c}{Test Accuracy}          \\ \cline{2-4} 
                         & 30\% dropped & 50\% dropped & 70\% dropped \\ \hline
SOFT-ELEM   & 0.980        & 0.976        & 0.963        \\
HARD-ELEM   & 0.796        & 0.734        & 0.772        \\
HARD-TIME   & 0.675        & 0.637        & 0.642        \\
STE-ELEM    & 0.942        & 0.945        & 0.956        \\
STE-TIME    & 0.637        & 0.659        & 0.679        \\ \hline
\textbf{SOFT-TIME} & \textbf{0.992}        & \textbf{0.989}        & \textbf{0.988}        \\ \hline
\end{tabular}
\end{table}

% \usepackage{multirow}
\begin{table}[!t]
\setlength{\tabcolsep}{2pt}
\centering
\caption{Ablation Study on PhysioNet Sepsis}\label{tbl:abl_sepsis}
\begin{tabular}{ccc}
\hline
\multicolumn{1}{c}{\multirow{2}{*}{Model}} & \multicolumn{2}{c}{Test AUCROC}                    \\ \cline{2-3} 
\multicolumn{1}{c}{}                       & \multicolumn{1}{c}{OI} & \multicolumn{1}{c}{No OI} \\ \hline
SOFT-TIME       & 0.891 & 0.831 \\
HARD-ELEM       & 0.869 & 0.847 \\
HARD-TIME       & 0.498 & 0.427 \\
STE-ELEM        & 0.702 & 0.637 \\
STE-TIME        & 0.597 & 0.551 \\ \hline
\textbf{SOFT-ELEM}  & \textbf{0.900}& \textbf{0.823}\\ \hline
\end{tabular}
\end{table}

\begin{table}[t]
\begin{minipage}{.48\linewidth}
\setlength{\tabcolsep}{1pt}
\scriptsize
\centering
\caption{Ablation Study On Regular Character Trajectories}\label{tbl:abl_charactertrajectory}
\begin{tabular}{cc}
\hline
Model & Test Accuracy \\ 
\hline
SOFT-ELEM       & 0.981         \\
HARD-ELEM       & 0.792        \\
HARD-TIME       & 0.604         \\
STE-ELEM        & 0.952         \\
STE-TIME        & 0.623         \\
\hline
\textbf{SOFT-TIME}   & \textbf{0.992}  \\
\hline
\end{tabular}
\end{minipage} \hfill
\begin{minipage}{.48\linewidth}
\setlength{\tabcolsep}{1pt}
\scriptsize
\centering
\caption{Ablation Study on Regular Google Stock}\label{tbl:abl_stock}
\begin{tabular}{cc}
\hline
Model & Test MSE \\ 
\hline
SOFT-ELEM           & 0.001      \\
SOFT-TIME           & 0.002      \\
HARD-ELEM           & 0.001      \\
HARD-TIME           & 0.047      \\
STE-TIME            & 0.022      \\
\hline
\textbf{STE-ELEM}   & \textbf{0.001}    \\
\hline
\end{tabular}
\end{minipage}
\end{table}

\begin{table}[!t]
\setlength{\tabcolsep}{2pt}
\centering
\caption{Ablation Study on Irregular Google Stock}\label{tbl:abl_google2}
\begin{tabular}{cccc}
\hline
\multirow{2}{*}{Model}   & \multicolumn{3}{c}{Test MSE}          \\ \cline{2-4} 
                         & 30\% dropped & 50\% dropped & 70\% dropped \\ \hline
SOFT-ELEM   & 0.001        & 0.002        & 0.001        \\
SOFT-TIME   & 0.001        & 0.002        & 0.002        \\
HARD-ELEM   & 0.002        & 0.001        & 0.002        \\
HARD-TIME   & 0.051        & 0.048        & 0.048        \\
STE-TIME    & 0.021        & 0.022        & 0.023        \\ 
\hline
\textbf{STE-ELEM} & \textbf{0.001}& \textbf{0.001} & \textbf{0.001} \\ \hline
\end{tabular}
\end{table}

\begin{figure}[t]
    \centering
    \subfigure[Patient \#1]{\includegraphics[width=0.23\textwidth]{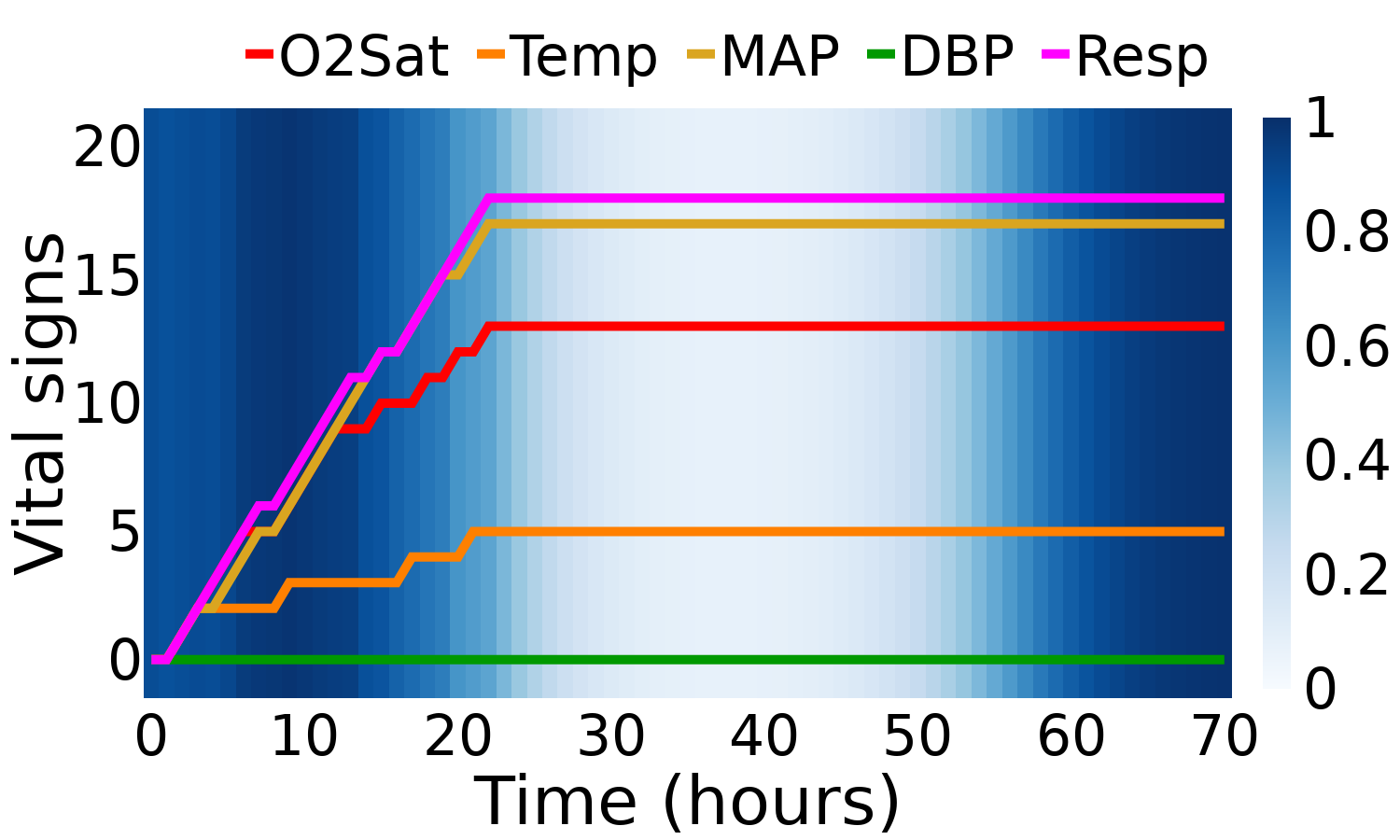}} \hfill
    \subfigure[Patient \#2]{\includegraphics[width=0.23\textwidth]{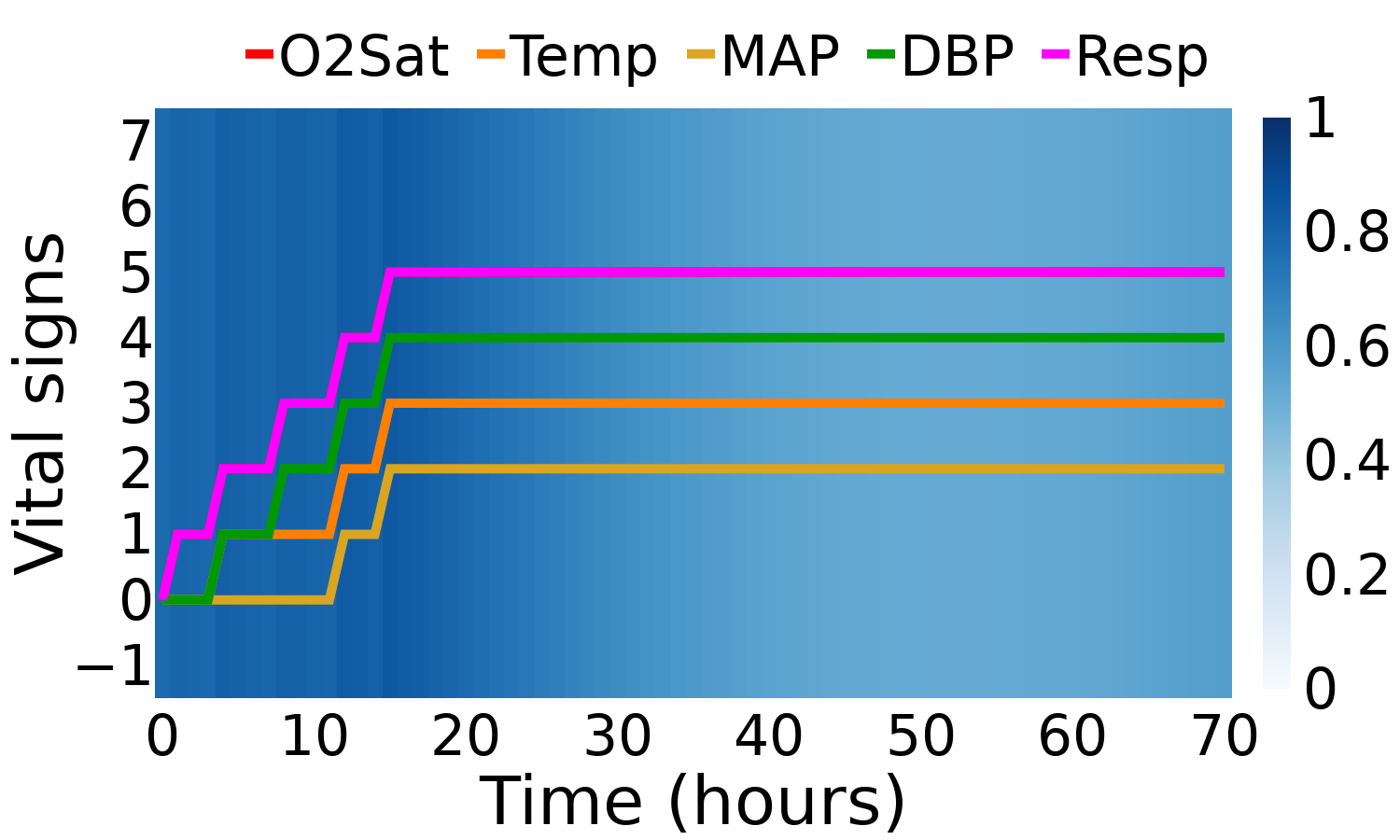}}\\
    \subfigure[Patient \#3]{\includegraphics[width=0.23\textwidth]{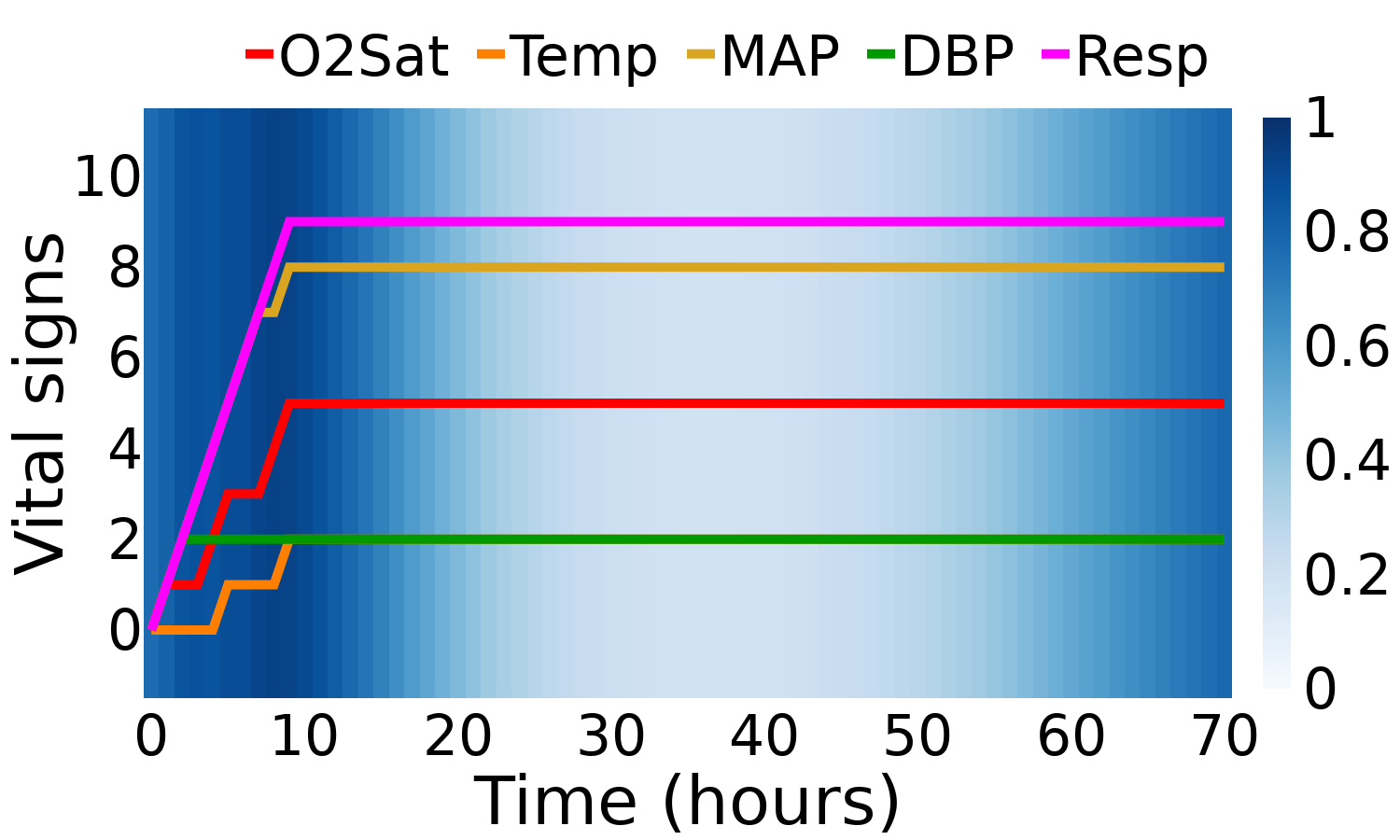}}\hfill
    \subfigure[Patient \#4]{\includegraphics[width=0.23\textwidth]{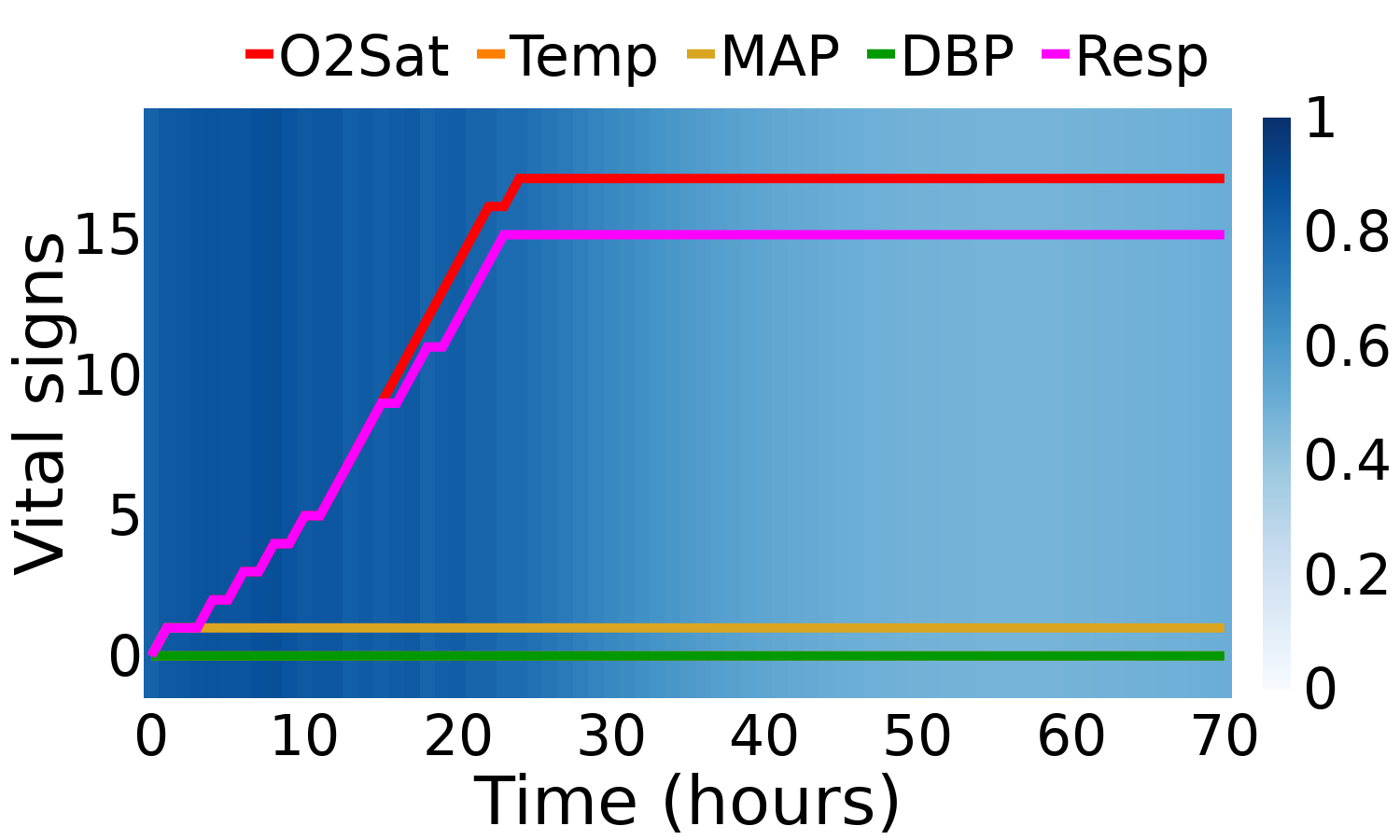}}\\
    
    \caption{The soft attention visualization in PhysioNet Sepsis. Blue means strong attention and white means weak attention. Note that the bottom NCDE gives strong attention when there are non-trivial changes on the curves.} 
    \label{fig:attn1}
\end{figure}

\begin{figure}[t]
    \centering
    \subfigure[Open price]{\includegraphics[width=0.23\textwidth]{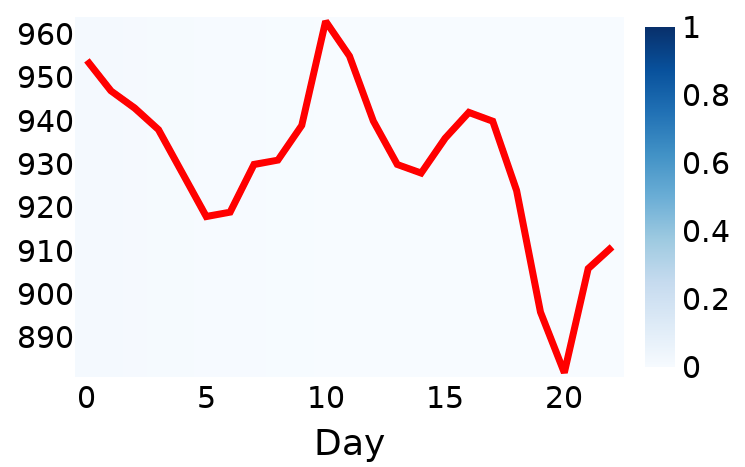}} \hfill
    \subfigure[High price]{\includegraphics[width=0.23\textwidth]{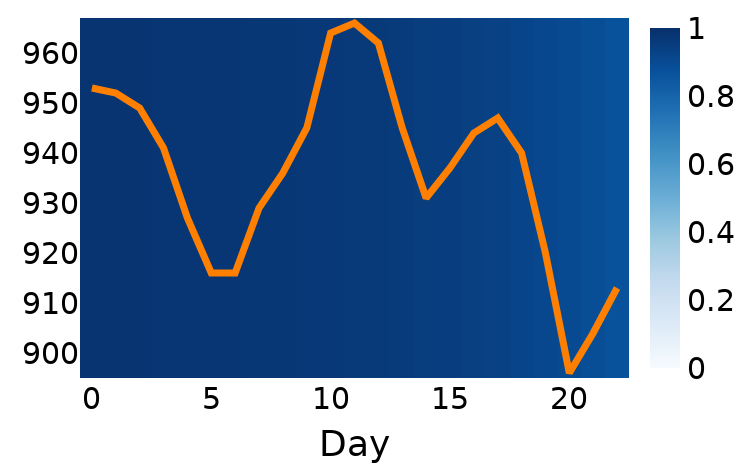}}\\
    \subfigure[Low price]{\includegraphics[width=0.23\textwidth]{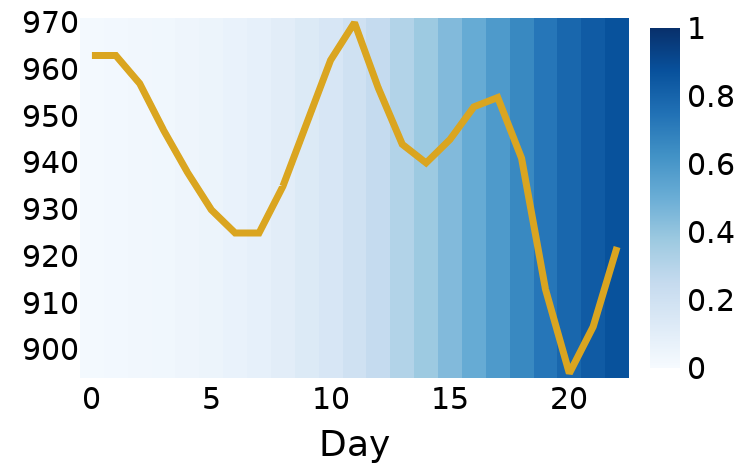}}\hfill
    \subfigure[Close price]{\includegraphics[width=0.23\textwidth]{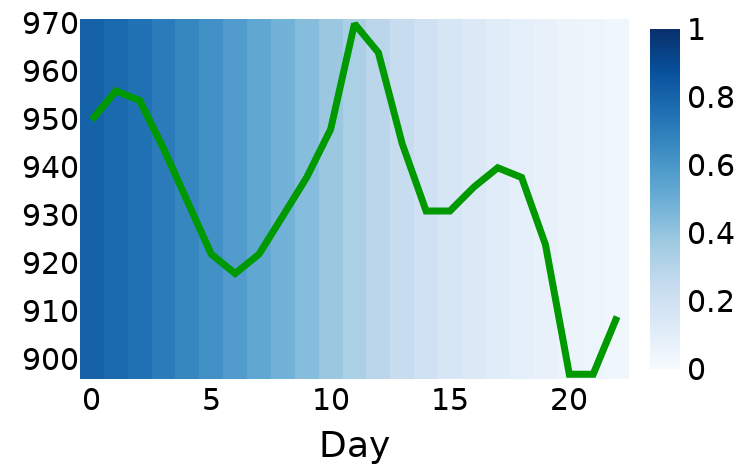}}\\
    \subfigure[Adjusted close price]{\includegraphics[width=0.23\textwidth]{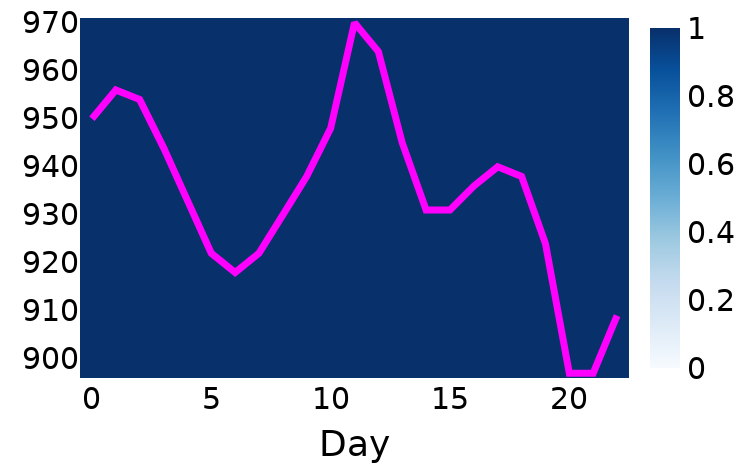}}\hfill
    \subfigure[Volume]{\includegraphics[width=0.23\textwidth]{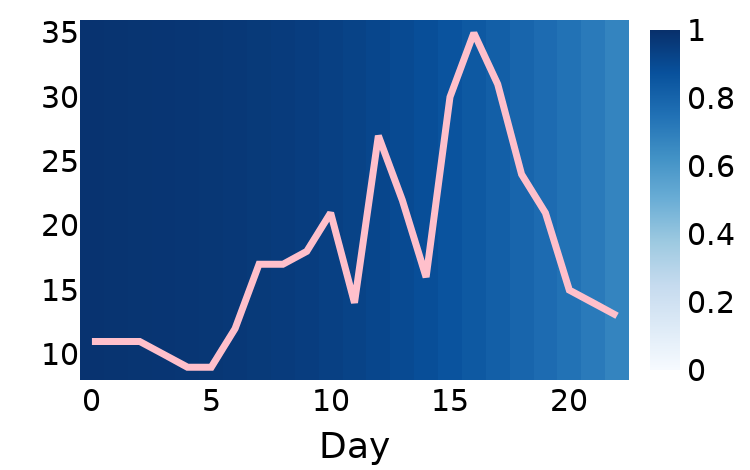}}\\
    
    \caption{The soft attention visualization in Google Stock. Blue means strong attention and white means weak attention. Note that the open price has weak attention always. This is because it shows high correlations to other values. Our attention model strategically ignores the open price.} 
    \label{fig:attn2}
\end{figure}

\medskip\noindent{\textbf{Google Stock (Regular): } The results of the regular time-series forecasting with Google Stock is in Table~\ref{tbl:google0}. Surprisingly, RNN-based models show small errors. They outperform many NODE and NCDE-based models. ACE-NODE shows the smallest mean error along with the smallest standard deviation among all baselines, which may show the efficacy of the attention mechanism for NODEs. However, ANCDE shows the smallest mean error with the smallest standard deviation for this dataset. Interestingly, NCDE shows the worst performance but our attention mechanism significantly stabilizes the performance, i.e., NCDE shows an MSE of 0.304 while ANCDE reduces MSE to 0.001.
}

\medskip\noindent{\textbf{Google Stock (Irregular): } Table~\ref{tbl:google1} shows irregular time-series forecasting results. The overall tendency is similar to that of the regular time-series experiment. GRU-based methods show reliable performance and NCDE performs poorly. However, ANCDE successfully stabilizes the performance again.
}

\subsection{Ablation Studies}
We additionally conduct ablation studies w.r.t. the six attention types in Tables~\ref{tbl:abl_charactertrajectory2} to~\ref{tbl:abl_google2}. We omit the ablation studies for regular time-series data due to space constraints.

In Table~\ref{tbl:abl_charactertrajectory2}, the soft time-wise attention shows the highest accuracy for the irregular time-series classification. For this dataset, the soft time-wise attention is preferred.

For PhysioNet Sepsis, however, the soft element-wise attention shows the highest AUCROC score as shown in Table~\ref{tbl:abl_sepsis}. In this dataset, the performance gap between attention types is very large, e.g., an AUCROC of 0.498 by the hard time-wise attention vs. 0.900 by the soft element-wise attention.

For Google Stock, however, the element-wise attention with STE shows the smallest errors in general. The hard time-wise attention shows the worst-case error, which is two orders of magnitude larger than that of the element-wise attention with STE.

\subsection{Visualization of Attention and Training Process}

Figs.~\ref{fig:attn1} and~\ref{fig:attn2} show how our proposed attention mechanism works. Values highlighted in blue indicate crucial information. In PhysioNet Sepsis, for instance, strong attention scores are given when there are non-trivial changes on the curves.

In Google Stock, the attention mechanism works in a more aggressive fashion. 
%It strategically ignores the entire (or partial) portion of the open/low/close prices for their high correlations to other values --- note that there exist many white regions in those three price figures. Since they can be readily reconstructed from others, our proposed method ignores them. Similar attention patterns are consistently observed in this dataset.
It entirely (or partially) ignores the open/low/close prices --- note that these three price figures are mostly white. Since future prices can be readily reconstructed from other parts of the data, our proposed method ignores these data values. Similar attention patterns are consistently observed in this dataset.

% \begin{figure}
%     \centering
%     \includegraphics[width=1.1\columnwidth]{images/stockattention.png}
%     \caption{The hard attention visualization of Google Stock dataset. The area highlighted in blue (resp. white) means an attention of 1 (resp. 0).}
%     \label{fig:attn2}
% \end{figure}

\section{Conclusions}
Recently, differential equation-based methods have been widely used for time-series classification and forecasting. Ever since NODEs were first introduced, many advanced concepts building upon NODEs have been proposed. However, NCDEs are still a rather unexplored field. To this end, we propose ANCDE, a novel NCDE architecture that incorporates the concept of attention. ANCDE consists of two NCDEs, where the bottom NCDE provides the top NCDE with attention scores, and the top NCDE utilizes these scores to perform a downstream machine learning task on time-series data. Our experiments with three real-world time-series datasets and ten baseline methods successfully prove the efficacy of the proposed concept. In addition, our visualization of attention intuitively delivers how our proposed method works.

% In the future, we will further extend our model with supervised attention. Our proposed attention mechanism significantly improves NCDEs. However, the supervised attention, which uses a small set of training data to supervise attention, was very successful in several domains~\cite{liu2016neural, liu2017exploiting, tang2019progressive}. Therefore, our next step to improve ANCDEs can be supervised attention.

\section*{Acknowledgement}
Noseong Park is the corresponding author. This work was supported by the Yonsei University Research Fund of 2021, and the Institute of Information \& Communications Technology Planning \& Evaluation (IITP) grant funded by the Korean government (MSIT) (No. 2020-0-01361, Artificial Intelligence Graduate School Program (Yonsei University), and No. 2021-0-00155, Context and Activity Analysis-based Solution for Safe Childcare).

\bibliographystyle{IEEEtran}
\bibliography{ref}

% \section*{Acknowledgment}

% The preferred spelling of the word ``acknowledgment'' in America is without 
% an ``e'' after the ``g''. Avoid the stilted expression ``one of us (R. B. 
% G.) thanks $\ldots$''. Instead, try ``R. B. G. thanks$\ldots$''. Put sponsor 
% acknowledgments in the unnumbered footnote on the first page.

\end{document}